\documentclass{article}

\usepackage[utf8]{inputenc} 
\usepackage[T1]{fontenc}    
\usepackage{hyperref}       
\usepackage{url}            
\usepackage{booktabs}       
\usepackage{amsfonts}       
\usepackage{nicefrac}       
\usepackage{microtype}      

\usepackage{lettrine}
\usepackage{csquotes}

\usepackage{comment}
\usepackage{subfigure}
\usepackage{color, xcolor}

\RequirePackage{mathtools}

\usepackage{bm}
\usepackage{natbib}
\newcommand{\lp}{\left(}
\newcommand{\rp}{\right)}
\newcommand{\lb}{\left\{}
\newcommand{\rb}{\right\}}
\newcommand{\lab}{\left|}
\newcommand{\rab}{\right|}

\newcommand{\ra}{\rightarrow }

\newcommand{\real}{\mathbb{R}}

\newcommand{\Rp}{\mathbb{R}^{p}}

\newcommand{\eop}{\hfill $\Box$}

\newcommand{\limt}{\lim_{t \rightarrow \infty}}

\newcounter{Lem}[section]
\renewcommand{\theLem}{\thesection.\arabic{Lem}}
\newenvironment{Lemma}{{\noindent\bf Lemma \stepcounter{Lem}\theLem.}\hspace{4pt}}

\newcounter{Exam}[section]
\renewcommand{\theExam}{\thesection.\arabic{Exam}}
\newenvironment{Example}{{\noindent\bf Example \stepcounter{Exam}\theExam.}\hspace{4pt}}

\newcounter{Thm}[section]
\renewcommand{\theThm}{\thesection.\arabic{Thm}}
\newenvironment{Theorem}{{\noindent\bf Theorem \stepcounter{Thm}\theThm.}\hspace{4pt}}

\newcounter{Prop}[section]
\renewcommand{\theProp}{\thesection.\arabic{Prop}}
\newenvironment{Proposition}{{\noindent\bf Proposition \stepcounter{Prop}\theProp.}\hspace{4pt}}

\usepackage{cleveref}
\crefformat{equation}{(#2#1#3)}
\crefrangeformat{equation}{(#3#1#4) to~(#5#2#6)}

\usepackage[english]{babel}

\title{The Implicit Bias of AdaGrad\\ on Separable Data}

\author{\textsc{Qian Qian}
\\[2mm]
Department of Statistics, The Ohio State University
\\
Columbus, OH 43210, USA
\\
\texttt{qian.216@osu.edu}
\\[6mm]
\textsc{Xiaoyuan Qian}
\\[2mm]
School of Mathematical Sciences, Dalian University of Technology
\\
Dalian, Liaoning 116024, China
\\
\texttt{xyqian@dlut.edu.cn}
}

\begin{document}

\maketitle

\begin{abstract}
  We study the implicit bias of AdaGrad on separable linear classification problems.
  We show that AdaGrad converges to a direction that can be  characterized as the solution of a quadratic optimization problem with the same feasible set as the hard SVM problem.
  We also give a discussion about how different choices of the hyperparameters of AdaGrad might impact this direction.
  This provides a deeper understanding of why adaptive methods do not seem to have the generalization ability as good as gradient descent does in practice.
\end{abstract}

\section{Introduction}

In recent years, implicit regularization from various optimization algorithms plays a crucial role in the generalizatiion abilities in training deep neural networks \citep{neyshabur15a, neyshabur15b, keskar16, neyshabur17, zhang17}. For example, in underdetermined problems where the number of parameters is larger than the number of training examples, many global optimum fail to exhibit good generalization properties, however, a specific optimization algorithm (such as gradient descent) does converge to a particular optimum that generalize well, although no explicit regularization is enforced when training the model. In other words, the optimization technique itself "biases" towards a certain model in an implicit way (\citep{soudry18}). This motivates a line of works to investigate the implicit biases of various algorithms \citep{telgarsky13, soudry18, gunasekar17, gunasekar18a, gunasekar18b}.

The choice of algorithms would affect the implicit regularization introduced in the learned models. In underdetermined least squares problems, where the minimizers are finite, we know that gradient descent yields the minimum $L_2$ norm solution, whereas coordinate descent might give a different solution. Another example is logisitic regression with separable data. While gradient descent converges in the direction of the hard margin support vector machine solution \citep{soudry18}, coordinate descent converges to the maximum $L_1$ margin solution \citep{telgarsky13,gunasekar18a}. Unlike the squared loss, the logistic loss does not admit a finite global minimizer on separable data: the iterates will diverge in order to drive the loss to zero. As a result, instead of characterizing the convergence of the iterates $\bm{w}(t)$, it is the asymptotic direction of these iterates i.e., $\lim_{t \rightarrow \infty} \bm{w}(t)/{\|\bm{w}(t)\|}$ that is important and therefore has been characterized (\citep{soudry18, gunasekar18b}).

Morevoer, it has attracted much attention
that different adaptive methods of gradient descent and hyperparameters of an adaptive method
exhibit different biases, thus leading to different generalization performance in deep learning \citep{neyshabur15a, keskar16, wilson17, hoffer17}.
Among those findings is that the vanilla SGD algorithm demonstrates better generalization than its adaptive variants \citep{wilson17}, such as AdaGrad \citep{duchi10} and Adam \citep{kingma15}.
Therefore it is important to precisly characterize how different adaptive methods induce difference biases. A natural question to ask is: can we explain this observation by characterizing the implicit bias of AdaGrad, which is a paradigm of adaptive methods, in a binary classification setting with separable data using logistic regression? And how does the implicit bias depend on the choice of the hyperparameters of this specific algorithm, such as initialization, step sizes, etc?

\subsection{Our Contribution}

In this work we study Adagrad applied to logisitc regression with separable data. Our contribution is three-fold as listed as follows.

\begin{itemize}
  \item We prove that the directions of AdaGrad iterates, with a constant step size sufficiently small, always converge.
  \item We formulate the asymptotic direction as the solution of a quadratic optimization problem. This achieves a theoretical characterization of the implicit bias of AdaGrad, which also provides insights about why and how the factors involved, such as certain intrinsic proporties of the dataset, the initialization and the learning rate, affect the impicit bias.
  \item We introduce a novel approach to study the bias of AdaGrad. It is mainly based on a geometric estimation on the directions of the updates, which doesn't depend on any calculation on the convergence rates.
\end{itemize}

\subsection{Paper Organization}

This paper is organized as follows.
In Section 2 we explain our problem setup.
The main theory is developed in Section 3, including convergence of the adaptive learning rates of AdaGrad, existence of the asymptotic direction of AdaGrad iterates, and relations between the asymptotic directions of Adagrad and gradient descent iterates.
We conclude our  paper in Section 4 with a review of our results and some questions left to future research.

\section{ Problem Setup }

Let
$ \lb \lp\bm{x}_n, y_{n}\rp :\ n =1,\cdots, N \rb $
be a training dataset with features $ \bm{x}_n\in\Rp $ and labels $ y_{n}\in \lb -1, 1\rb\,.$
Consider learning the logistic regression model over the empirical loss:
\begin{equation*}
    \mathcal{L}(\bm{w}) = \sum_{n=1}^{N} l\lp y_n \bm{w}^T\bm{x}_{n} \rp \,,\ \ \ \bm{w} \in \real^p \,,
\end{equation*}
where $ l: \Rp \ra \real\,.$
We focus on the following case, same as proposed in \citep{soudry18}:

{\bf Assumption 1.} There exists a vector $ \bm{w}_{\ast} $ such that
$ \ y_{n}\bm{w}_{\ast}^T\bm{x}_{n} > 0 $  for all $ n $.

{\bf Assumption 2.} $ l(u) $ is continuously differentiable, $\beta-$smooth, and strictly decreasing to zero.

{\bf Assumption 3.}  There exist positive constants $ a, b, c, $ and $d$ such that
\begin{equation*}
   \lab l'(u) + c e^{-au} \rab \leq e^{-(a+b)u}\,,\ \ \text{for}\ u > d \,.
\end{equation*}

It is easy to see that the exponential loss $ l(u) = e^{-u} $ and the logistic loss $ l(u) = \log\lp 1+e^{-u} \rp $
both meet these assumptions.

Given two hyperparameters $ \epsilon\,, \ \eta > 0\, $ and an initial point $ \bm{w}(0) \in \Rp \,, $
we consider the diagonal AdaGrad iterates
\begin{equation}\label{1}
    \bm{w}(t+1) = \bm{w}(t) -\eta \bm{h}(t) \odot \bm{g}(t)\,, \ \ \ t =0, 1, 2, \cdots\,,
\end{equation}
where
$$
    \bm{g}(t) =\lp g_{1}(t),\cdots, g_{p}(t) \rp  \,,
$$
$$
      g_{i}(t) =  \frac{\partial \mathcal{L}}{\partial w_{i}} ( \bm{w}(t) ) \,,
$$
$$
    \bm{h}(t) =\lp h_{1}(t),\cdots, h_{p}(t) \rp  \,,
$$
$$
      h_{i}(t) =  \frac{1}{ \sqrt{ g_{i}(0)^{2} +\cdots + g_{i}(t)^{2} + \epsilon} } \,, \ \ \ i = 1, \cdots, p \,,
$$
and $ \odot $ is the element-wise multiplication of two vectors, e.g.
$$
    \bm{a}\odot \bm{b} = \lp a_{1}b_{1},\cdots, a_{p}b_{p} \rp^T
$$
for $ \bm{a} = \lp a_{1},\cdots, a_{p} \rp^T,\  \bm{b} = \lp b_{1},\cdots, b_{p} \rp^T .$

To analyze the convergence of the algorithm, we put an additional restriction on the hyperparameter $\ \eta\ $.

{\bf Assumption 4.}  The hyperparameter $\ \eta\ $ is not too large; specifically,
\begin{equation}\label{2}
   \eta < \frac{2\ \min_{i\in\lb 1,\cdots, p\rb}\sqrt{g_{i}(0)^2+ \epsilon}}{\beta} .
\end{equation}

We are interested in the asymptotic behavior of the AdaGrad iteration scheme in \eqref{1}. The main problem is:
does there exists some vector $ \bm{w}_{A} $ such that
$$
     \lim_{t\rightarrow \infty} \left. \bm{w}(t) \right/ \| \bm{w}(t) \|  = \bm{w}_{A} \,?
$$
We will provide an affirmative answer to this question in the following section.

\section{ The Asymptotic Direction of AdaGrad Iterates}
%

\subsection{ Convergence of the Adaptive Learning Rates }

We first provide some elementary facts about AdaGrad iterates \eqref{1} with all assumptions (1-4) proposed in Section 2.

\bigskip

\begin{Lemma}
$ \ \mathcal{L}\lp \bm{w}(t+1) \rp  <  \mathcal{L}\lp \bm{w}(t) \rp \ \ (\ t =0, 1, \cdots)\,. $
\end{Lemma}

\bigskip

\begin{Lemma}\label{finite_sum_gt}
$ \
  \sum_{t=0}^{\infty}\| \bm{g}(t) \|^2 < \infty \,. 
$
\end{Lemma}

We notice that \cite{gunasekar18a} showed a similar result (Lemma 6, in Section 3.3 of their work) for exponential loss only, under slightly different assumptions.
However, their approach depends on some specific properties of the exponential function, and thus cannot be extended to Lemma 3.2 in a trivial manner.

\bigskip

\begin{Lemma}
The following statements hold:

(i) $\ \| \bm{g}(t)\| \ra 0\ \ (t\ra\infty) $.

(ii) $\ \| \bm{w}(t)\| \ra \infty\ \ (t\ra\infty) $.

(iii) $\ \mathcal{L}(\bm{w}(t)) \ra 0 \ \ (t\ra\infty) $.

(iv) $\ \forall n\,,\ \ \lim_{t\ra\infty} \bm{w}(t)^T\bm{x}_n = \infty $.

(v) $\ \exists t_0\,,\ \ \forall\  t> t_0\,,\ \ \bm{w}(t)^T \bm{x}_n > 0$.
\end{Lemma}

\bigskip


\begin{Theorem}
The sequence $ \lb \bm{h}(t) \rb_{t=0}^{\infty} $ converges as $ t\ra \infty $ to a vector
$$
    \bm{h}_{\infty} = \lp h_{\infty, 1},\cdots, h_{\infty, p} \rp
$$
satisfying $\  h_{\infty, i} > 0\ \ (i = 1,\cdots, p) \,.$
\end{Theorem}

\subsection{ Convergence of the Directions of AdaGrad Iterates }

In the remainder of the paper we denote
$\ \bm{h}_{\infty} = \lim_{t\ra\infty}\bm{h}(t)\,$ and $\ \bm{\xi}_{n} = \bm{h}_{\infty}^{1/2}\odot \bm{x}_{n}\ \ (n=1,\cdots, N)\,.$
Since, by Theorem 3.1, the components of $\ \bm{h}_{\infty} \,$ have a positive lower bound,
we can define
$$
    \bm{\beta}(t) = \bm{h}_{\infty}^{-1}\odot \bm{h}(t)\ \ (t=0, 1, 2, \cdots)\,.
$$
Here the squared root and the inverse of vectors are defined element-wise.
We call the function
\begin{equation*}
  \mathcal{L}_{ind}: \Rp \ra \real \,,\ \ \ \mathcal{L}_{ind}(\bm{v}) = \sum_{n=1}^{N} l \lp \bm{v}^{T} \bm{\xi}_{n}\rp
\end{equation*}
the \emph{induced loss} with respect to AdaGrad \eqref{1}.
Note that
\begin{align*}
   \mathcal{L}(\bm{w})
   &= \sum_{n=1}^{N} l \lp \bm{w}^{T} \bm{x}_{n} \rp  \nonumber = \sum_{n=1}^{N} l \lp \lp \bm{h}_{\infty}^{-1/2}\odot \bm{w}\rp^{T}\lp \bm{h}_{\infty}^{1/2}\odot \bm{x}_{n}\rp \rp   \\
   &= \sum_{n=1}^{N} l \lp \lp \bm{h}_{\infty}^{-1/2}\odot \bm{w}\rp^{T}\bm{\xi}_{n} \rp = \mathcal{L}_{ind}\lp \bm{h}_{\infty}^{-1/2}\odot \bm{w} \rp  \,.
\end{align*}
Thus if we set
\begin{equation}\label{vnw}
  \bm{v}(t) = \bm{h}_{\infty}^{-1/2}\odot \bm{w}(t)\ \ (t = 0,1,2,\cdots) \,,
\end{equation}
then
$\ \bm{v}(0)  = \bm{h}_{\infty}^{-1/2} \odot \bm{w}(0)\,,$
and
\begin{align*}
   \bm{h}_{\infty}^{1/2}\odot \bm{v}(t+1)
   &= \bm{w}(t+1) =  \bm{w}(t) -\eta \bm{h}(t) \odot \nabla \mathcal{L} \lp \bm{w}(t) \rp \\
   &=  \bm{h}_{\infty}^{1/2}\odot \bm{v}(t) -\eta \bm{h}(t)\odot \bm{h}_{\infty}^{-1/2}\odot\nabla \mathcal{L} \lp \bm{h}_{\infty}^{1/2}\odot \bm{v}(t) \rp \\
   &=  \bm{h}_{\infty}^{1/2}\odot \bm{v}(t) -\eta \bm{h}(t)\odot\bm{h}_{\infty}^{-1/2}\odot \nabla \mathcal{L}_{ind} \lp \bm{h}_{\infty}^{-1/2}\odot \lp \bm{h}_{\infty}^{1/2}\odot \bm{v}(t) \rp \rp \\
   &= \bm{h}_{\infty}^{1/2}\odot \lp \bm{v}(t) -\eta \bm{\beta}(t)\odot\nabla \mathcal{L}_{ind} \lp \bm{v}(t)\rp \rp   \,,
\end{align*}
or
\begin{equation}\label{7}
  \bm{v}(t+1) = \bm{v}(t) - \eta \bm{\beta}(t) \odot \nabla \mathcal{L}_{ind}(\bm{v}(t))\ \ \ (t= 0,1,\cdots) \,.
\end{equation}
We refer to \eqref{7} as the \emph{induced form} of AdaGrad \eqref{1}.

The following result for the induced form is a simple corollary of Lemma 3.3.

\bigskip

\begin{Lemma}
The following statements hold:

(i) $\ \| \nabla\mathcal{L}_{ind}(t)\| \ra 0\ \ (t\ra\infty) $.

(ii) $\ \| \bm{v}(t)\| \ra \infty\ \ (t\ra\infty) $.

(iii) $\ \mathcal{L}_{ind}(\bm{v}(t)) \ra 0 \ \ (t\ra\infty) $.

(iv) $\ \forall n\,,\ \ \lim_{t\ra\infty} \bm{v}(t)^T\bm{\xi}_n = \infty $.

(v) $\ \exists t_0\,,\ \ \forall\  t> t_0\,,\ \ \bm{v}(t)^T \bm{\xi}_n > 0$.
\end{Lemma}

For the induced loss $\ \mathcal{L}_{ind}$, Consider GD iterates
\begin{align}\label{9}
  \bm{u}(t+1) & = \bm{u}(t) - \eta \nabla \mathcal{L}_{ind}(\bm{u}(t))\ \ \ (t= 0,1,\cdots)\,.
\end{align}
According to Theorem 3 in Soudry et.al.(2018), we have
$$
     \lim_{t\ra \infty} \frac{ \bm{u}(t) }{\|  \bm{u}(t) \|} = \frac{\widehat{\bm{u}}}{\|\widehat{\bm{u}}\|} \,,
$$
where
\begin{equation*}
    \widehat{\bm{u}} = \underset{\bm{u}^T \bm{\xi}_{n}\geq 1,\, \forall n}{\arg\min} \| \bm{u} \|^{2} .
\end{equation*}
Noting that $\ \bm{\beta}(t) \ra \bm{1} \ \ (t\ra\infty) \,$
we can obtain GD iterates \eqref{9} by taking the limit of $ \bm{\beta}(t) $ in \eqref{7}.
Therefore it is reasonable to expect that these two iterative processes have similar asymptotic behaviors, especially a common limiting direction.

Different from the case of GD method discussed in \citep{soudry18}, however, it is difficult to obtain an effective estimation about the convergence rate of $ \bm{w}(t) $.
Instead, we introduce an orthogonal decomposition approach to obtain the asymptotic direction of the original Adagrad process \eqref{1}.

In the remainder of the paper, we denote by $ P $ the projection onto the $1-$dimensional subspace spanned by $ \widehat{\bm{u}} $,
and $ Q $ the projection onto the orthogonal complement.
Without any loss of generality we may assume
$\ \|\widehat{\bm{u}}\| = 1 \,.$
Thus we have
the orthogonal decomposition
\begin{equation*}
    \bm{v} = P\bm{v} + Q\bm{v}\ \ \ (\bm{v} \in \Rp)\,,
\end{equation*}
where
$ P\bm{v} = \|P\bm{v}\| \widehat{\bm{u}} = \lp \bm{v}^{T}\widehat{\bm{u}} \rp \widehat{\bm{u}}.
$
Moreover, we denote
\begin{equation}\label{11}
  \bm{\delta}(t) = - \eta \nabla \mathcal{L}_{ind}\lp \bm{v}(t) \rp , \ \ \ \
   \bm{d}(t) = \bm{\beta}(t)\odot \bm{\delta}(t) .
\end{equation}
Using this notation we can rewrite the iteration scheme \eqref{7} as
$$
   \bm{v}(t+1) = \bm{v}(t) + \bm{d}(t) \quad  (t= 0,1,\cdots) .
$$
By reformulating \eqref{11} as
$$
    \bm{d}(t) = \bm{\delta}(t) +\lp \bm{\beta}(t) - \bm{1}\rp \odot \bm{\delta}(t) \,,
$$
where $ \bm{\beta}(t) - \bm{1} \ra \bm{0} $ as $ t\ra\infty $,
we regard $ \bm{\delta}(t) $ as the decisive part of $ \bm{d}(t) $
and acquire properties of $ \bm{d}(t) $
through exploring analogues of $ \bm{\delta}(t) $.

First, we can show a basic estimation:
\begin{equation*}
    \bm{\delta}(t)^T\widehat{\bm{u}} = \| P \bm{\delta}(t)\| \geq \frac{\|\bm{\delta}(t) \|}{\underset{n}{\max}\ \|\bm{\xi}_{n}\|}
    \ \ \ (t = 0,1,2\cdots)\,.
\end{equation*}
The projection properties of $ \bm{\delta}(t) $ is easily passed on to $ \bm{d}(t) $.
In fact,
for sufficiently large $\ t \,,$
\begin{equation}\label{14}
    \bm{d}(t)^{T}\widehat{\bm{u}} = \| P \bm{d}(t)\| \geq  \frac{ \|\bm{d}(t) \|}{4\,\underset{n}{\max}\ \|\bm{\xi}_{n}\|} \,,
\end{equation}
Inequality \eqref{14} provides a cumulative effect on the projection of $ \bm{v}(t) $
as $ t $ increases:
$$
    \| P \bm{v}(t)\| \geq \frac{\|\bm{v}(t) \|}{8\underset{n}{\max}\ \|\bm{\xi}_{n}\|}
    \,, \ \ \text{for sufficiently large $\ t $} \,.
$$

\medskip

The following lemma reveals a crucial characteristic of the iterative process \eqref{7}:
as $ t $ tends to infinity,
the contribution of $ \bm{\delta}(t)$ to the increment of the deviation from the direction of $ \widehat{\bm{u}} $ , compared to its contribution to the increment in the direction of $ \widehat{\bm{u}} $, becomes more and more insignificant.

\bigskip

\begin{Lemma}
Given $ \varepsilon > 0 \,.$
Let $ a,b,c \,$ be positive numbers as defined in Assumption 3 in Section 2.
If $\ \| Q\bm{v}(t) \| >2N(c+1)(ace\varepsilon)^{-1} \,,$
then for sufficiently large $\, t,$
\begin{equation*}
  Q\bm{v}(t) ^{T}\bm{\delta}(t) < \varepsilon \| Q\bm{v}(t)\|\| \bm{\delta}(t)\| \,.
\end{equation*}
\end{Lemma}

This property can be translated into a more convenient version for $ \bm{d}(t) $.

\bigskip

\begin{Lemma}
For any $ \varepsilon > 0 \,,$ there exist $ R > 0 $ such that for sufficiently large $\, t $ and  $\ \| Q\bm{v}(t)\| \geq R $,
\begin{equation*}
   \|  Q\bm{v}(t+1)\| - \|  Q\bm{v}(t)\| \leq \varepsilon \|\bm{d}(t)\|\,.
\end{equation*}
\end{Lemma}

Therefore, over a long period, the cumulative increment of $ \bm{v}(t) $ in the direction of $ \widehat{\bm{u}} $ will overwhelm the deviation from it, yielding the existence of an asymptotic direction for $ \bm{v}(t) $.

\bigskip

\begin{Lemma}
\begin{equation}\label{asymofv}
   \limt \frac{\bm{v}(t)}{\| \bm{v}(t)\|} =  \widehat{\bm{u}} \,.
\end{equation}
\end{Lemma}

\bigskip

By the relation \eqref{vnw} between $\ \bm{v}(t)\ $ and $\ \bm{w}(t),$
our main result directly follows from \eqref{asymofv}.

\medskip

\begin{Theorem}\label{main}
AdaGrad iterates \eqref{1} has an asymptotic direction:
\begin{equation*}
   \limt \frac{\bm{w}(t)}{\| \bm{w}(t)\|} = \frac{\widetilde{\bm{w}}}{\|\widetilde{\bm{w}}\|} ,
\end{equation*}
where
\begin{equation}\label{hop}
    \widetilde{\bm{w}} = \underset{\bm{w}^{T}\bm{x}_{n}\geq 1,\, \forall n }{ \arg\min } \left\|\frac{1}{\sqrt{\bm{h}_{\infty}}}\odot\bm{w}\right\|^2 \,.
\end{equation}
\end{Theorem}

\subsection{Factors Affecting the Asymptotic Direction }

Theorem 3.2 confirms that AdaGrad iterates \eqref{1}
have an asymptotic direction
$\ \widetilde{\bm{w}}/\| \widetilde{\bm{w}} \| \,,$
where $\ \widetilde{\bm{w}}\ $ is the solution to the optimization problem \eqref{hop}.
Since the objective function
$\ \left\|\bm{h}_{\infty}^{-1/2}\odot\bm{w}\right\|^2 $
is determined by the limit vector $\ \bm{h}_{\infty}\,,$
it is easy to see that the asymptotic direction may depend on the choices of
the dataset $ \lb \lp\bm{x}_n, y_{n}\rp \rb_{n=1}^{N} $ , the hyperparameters $ \epsilon\,, \ \eta\,,$ and the initial point $ \bm{w}(0) \,.$
In the following we will discuss this varied dependency in several respects.

\subsubsection{ Difference from the Asymptotic Direction of GD iterates }

When the classic gradient descent method
is applied to minimize the same loss,
it is known (see Theorem 3, \citep{soudry18}) that
GD iterates
\begin{equation}\label{29}
    \bm{w}_{G}(t+1) = \bm{w}_{G}(t) - \eta \nabla \mathcal{L}\lp \bm{w}_{G}(t)\rp\ \ \ ( t = 0,1,2,\cdots ) \,,
\end{equation}
have an asymptotic direction
$\ \left. \widehat{\bm{w}}\right/ \| \widehat{\bm{w}} \| $,
where $ \widehat{\bm{w}} $
is the solution of the hard max-margin SVM problem
\begin{equation}\label{27}
    \underset{ y_{n}\bm{w}^{T}\bm{x}_{n}\geq 1,\, \forall n }{ \arg\min } \|\bm{w}\|^2 \,.
\end{equation}
The two optimization problems \eqref{hop} and \eqref{27} have the same feasible set
$$
    \lb \bm{w}\in\Rp :\  y_{n}\bm{w}^{T}\bm{x}_{n}\geq 1 ,\ \text{for}\ n = 1,\cdots, N \rb \,,
$$
but they take on different objective functions.
It is natural to expect that their solutions $\ \widetilde{\bm{w}}\ $ and $\ \widehat{\bm{w}}\ $
yield different directions, as shown in the following toy example.

\bigskip

\begin{Example}
Let
$\ \bm{x}_{1} = \left(\cos \theta, \sin \theta \right)^T ,
$
$\ y_{1} = 1 ,$
$\ \bm{x}_{2} = -\bm{x}_{1} ,
$
$\ y_{2} = -1 ,$
and
$$
    \ \mathcal{L}(\bm{w}) =  e^{-\bm{w}^{T}\bm{x}_{1}}+ e^{\bm{w}^{T}\bm{x}_{2}} = 2 e^{-\bm{w}^{T}\bm{x}_{1}}  \,.
$$
Suppose $\ 0 < \theta < \pi/2\,.$
In this setting we simply have $\ \widehat{\bm{w}} = \bm{x}_{1} \,.$
Selecting $\ \bm{w}(0) = \left( a, b \right)^T $ and $\ \epsilon = 0 , $
we have
$$
    - \bm{g}(0) = 2 e^{-\bm{w}(0)^{T}\bm{x}_{1} }\bm{x}_{1}
               = 2 e^{-a\cos\theta - b\sin\theta } \left(\cos \theta, \sin \theta \right)^T ,
$$
$$
    \bm{h}(0) = \lp h_{1}(0), h_{2}(0) \rp^T
              = e^{a\cos\theta + b\sin\theta } \left( \frac{1}{\cos \theta},  \frac{1}{\sin \theta} \right)^T .
$$
In general we can show there is a sequence of positive numbers $\ p(t) $ such that
$$
   - \bm{g}(t) =  p(t) \left(\cos \theta, \sin \theta \right)^T ,
$$
and
\begin{equation*}
   \bm{h}_{\infty} = \limt \frac{1}{\sqrt{p(0)^2 + p(1)^2 + \cdots + p(t)^2}}\left( \frac{1}{\cos \theta},  \frac{1}{\sin \theta}  \right)^T = \frac{1}{\rho} \left( \frac{1}{\cos \theta},  \frac{1}{\sin \theta}  \right)^T .
\end{equation*}
Now
\begin{eqnarray*}
    \widetilde{\bm{w}} &=& \underset{\bm{w}^{T}\bm{x}_{1}\geq 1 }{ \arg\min } \left\|\bm{h}_{\infty}^{-1/2}\odot\bm{w}\right\|^2 =   \underset{\bm{w}^{T}\bm{x}_{1}\geq 1}{ \arg\min } \
    \rho \lp w_{1}^{2}\cos \theta + w_{2}^{2} \sin \theta\rp   \\
    &=&   \underset{\bm{w}^{T}\bm{x}_{1}\geq 1}{ \arg\min }
    \lp w_{1}^{2}\cos \theta + w_{2}^{2} \sin \theta\rp = \lp \frac{1}{\cos \theta + \sin \theta}, \frac{1}{\cos \theta + \sin \theta}\rp ,
\end{eqnarray*}
and we have
$\
  \widetilde{\bm{w}} /\|\widetilde{\bm{w}} \| = \left( \sqrt{2}/2, \sqrt{2}/2\right)^T \,.
$
Note that this direction is invariant when $ \theta $ ranges between $ 0 $ and $ \pi/2 $, i.e., irrelevant to $ \bm{x}_{1} $. These two directions coincide only when $\ \theta  = \pi/4 .$
\end{Example}

\subsubsection{ Sensitivity to Small Coordinate System Rotations }

If we consider the same setting as in Example 3.1, but taking $\theta \in \lp \pi/2, \pi \rp .$
Then the asymptotic direction  $ \widetilde{\bm{w}} /\|\widetilde{\bm{w}} \| $ will become
$\ \left( - \sqrt{2}/2, \sqrt{2}/2\right)^T.$
This implies, however, if $\ \bm{x}_{1}\ $ is close to the direction of $y-$axis,
then a small rotation of the coordinate system
may result in a large change of the asymptotic direction reaching a right angle, i.e.,
in this case the asymptotic direction is highly unstable even for a small perturbation of its $x-$coordinate.

\subsubsection{ Effects of the Initialization and Hyperparameter $\ \eta\ $ }

It is reasonable to believe that the asymptotic direction of AdaGrad depends on the initial conditions,
including initialization and step size (see Section 3.3, \citet{gunasekar18a}).
Theorem 3.2 yields a geometric interpretation for this dependency as shown in Figure 1,
where the red arrows indicate
$\ \bm{x}_{1} = \lp \cos \lp 3\pi/8 \rp, \sin \lp 3\pi /8\rp \rp $
and $\ \bm{x}_{2} = \lp \cos \lp 9\pi/20 \rp , \sin \lp 9\pi/20 \rp \rp $,
and the cyan arrow indicates the max-margin separator.
Since the isolines of the function
$ \left\|\bm{h}_{\infty}^{-1/2}\odot\bm{w}\right\|^2 $ are ellipses (drawn in green) centered at the origin,
the unique minimizer of the function in the feasible set (the grey shadowed area)
must be the tangency point (pointed at by the magenta arrow) between the tangent ellipse and the boundary of the feasible set.
If $\ \bm{h}_{\infty}\ $ varies, then the eccentricity of the tangent ellipses may change.
It makes the tangency point move along the boundary, indicating the change of the asymptotic direction.
\begin{figure}[h]
  \centering
  \includegraphics[width=6cm]{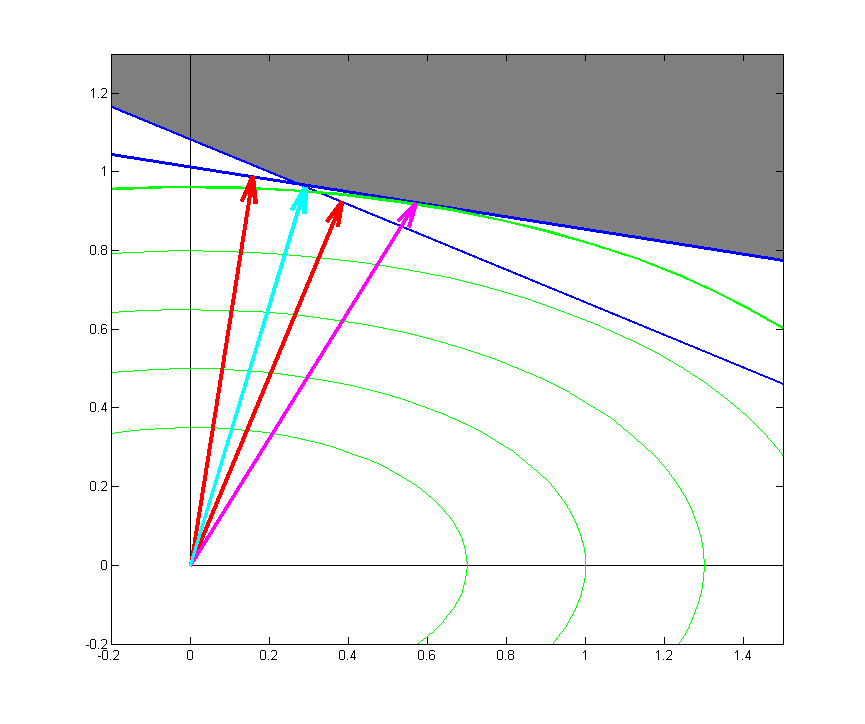}\\
  \caption{A case that the asymptotic directions of AdaGrad and GD are different.}
\end{figure}

\subsubsection{ Cases that the Asymptotic Direction is Stable }

Above we have observed that the asymptotic direction of AdaGrad iterates is very different from that of GD iterates.
We also find that there is a significant proportion of the cases that
their asymptotic directions are indeed equal.
Note that the asymptotic direction of GD iterates is robust with respect to different choices of initialization and learning rate $\ \eta\,,$
so is that of AdaGrad iterates in these cases.

\bigskip

\begin{Proposition}\label{positiveprb}
Suppose $\ N \geq p \,$ and $\ \bm{X} = \left[ \bm{x}_{1}, \cdots, \bm{x}_{N} \right]\in\real^{p\times N} \,$
is sampled from any absolutely continuous distribution.
Then with a positive probability the asymptotic directions of AdaGrad \eqref{1} and GD \eqref{29} are equal.
\end{Proposition}

\bigskip

\begin{Example}
Let $\ r_1,\ r_2 > 0$,
\begin{eqnarray*}
     \bm{x}_{1} &=& r_1\left(\cos \theta_1, \sin \theta_1 \right)^T\,,
     \ \ \ \frac{\pi}{2}\leq \theta_1 < \pi \,,    \\
     \bm{x}_{2} &=& r_2\left(\cos \theta_2, \sin \theta_2 \right)^T\,,
     \ \ \ \theta_{1} - \pi< \theta_2 \leq 0\,,
\end{eqnarray*}
and $\ \mathcal{L}(\bm{w}) =  l(\bm{w}^{T}\bm{x}_{1})+  l(\bm{w}^{T}\bm{x}_{2})\,.$
The system of equations
$$
   \bm{w}^T\bm{x}_{i} = 1\ \ \ (i=1,2)
$$
has a unique solution $\ (\alpha,\,\beta)^T $, where
\begin{eqnarray*}
   \alpha =  \frac{r_{2}^{-1}\sin \theta_{1} - r_{1}^{-1}\sin \theta_{2} }{\sin \lp \theta_{1} - \theta_{2} \rp } > 0 ,\ \ \
   \beta = \frac{r_{1}^{-1}\cos \theta_{2} - r_{2}^{-1}\cos \theta_{1} }{\sin \lp \theta_{1} - \theta_{2} \rp } > 0 \,.
\end{eqnarray*}
It is easy to check that if $\ \bm{w} = (w_1, w_2)^T $ satisfies
$\ \bm{w}^T\bm{x}_{i} \geq 1\ \ \ (i=1,2)\,, $
then $\ w_1 \geq \alpha\,,\ w_2 \geq \beta $.
Thus any quadratic form $\ b_1 w_1^2 + b_2 w_2^2 \ \ ( b_1,\, b_2 > 0)\ $
takes its minimum at $ (\alpha,\,\beta)^T $ over the feasible set
$ \lb  \bm{w}:\, \bm{w}^T\bm{x}_{i} \geq 1\ \ (i=1,2) \rb $.
Hence the asymptotic direction of AdaGrad \eqref{1}
applying to this problem is always equal to $ \left.(\alpha,\,\beta)^T \right/\|(\alpha,\,\beta)\| ,$
which is also the asymptotic direction of GD \eqref{29}.
\end{Example}

\begin{figure}[h]
  \centering
  \includegraphics[width=6cm]{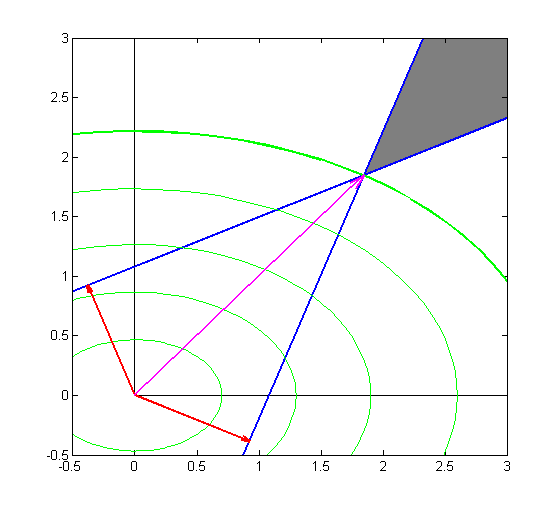}\\
  \caption{A case that the asymptotic directions of AdaGrad and GD are equal.}
\end{figure}

A geometric perspective of this example is given in Figure 2,
where the red arrows indicate
$$
 \bm{x}_{1} = \lp \cos \frac{5\pi}{8}, \sin \frac{5\pi}{8} \rp \ \ \text{and}\ \ \bm{x}_{2} = \lp \cos \frac{-\pi}{8} , \sin \frac{-\pi}{8} \rp,
$$
and the magenta arrow indicates $\ (\alpha,\,\beta)^T $.
It is easy to see that the isoline (the thick ellipse drawn in green) along which the function
$ \left\|\bm{h}_{\infty}^{-1/2}\odot\bm{w}\right\|^2 $ equals its minimum
must intersect with the feasible set (the grey shadowed area) at the corner $\ (\alpha,\,\beta)^T $, no matter what $ \bm{h}_{\infty} $  is.

\bigskip

The intuition shown in Example 3.2 can be extended to more general cases, as described in the following proposition.   
\bigskip

\begin{Proposition}\label{pprb}
Let
$ F =  \lb \bm{w}:\ \bm{w}^T\bm{x}_{n}\geq 1 \ (n=1,\cdots,N) \rb $
and let
$\ \bm{a} = \lp a_{1},\cdots, a_{p} \rp^T \ $ be a vector satisfying $\ a_{1}\cdots a_{p} \not= 0 \,.$
Suppose $ \bm{a} \in F\,,$
\begin{equation*}
   F \subset \lb \bm{a}+\bm{u}:\ \bm{u} = \lp u_{1},\cdots, u_{p} \rp^T\ \ \text{such that}\ \ a_{i}u_{i}\geq 0\ (i=1,\cdots,p) \rb\,,
\end{equation*}
and $\ \bm{b} = \lp b_{1},\cdots, b_{p} \rp^T \in\Rp \,.$
Then
\begin{equation*}
\underset{\bm{w}^{T}\bm{x}_{n}\geq 1,\, \forall n }{ \arg\min } \left\| \bm{b}\odot\bm{w}\right\|^2
= \underset{\bm{w}^{T}\bm{x}_{n}\geq 1,\, \forall n }{ \arg\min } \left\| \bm{w}\right\|^2 = \bm{a}\,,
\end{equation*}
and therefore the asymptotic directions of AdaGrad \eqref{1} and GD \eqref{29} are equal.
\end{Proposition}

\section{ Conclusion }


We proved that the basic diagonal AdaGrad, when minimizing a smooth monotone loss function
with an exponential tail, has an asymptotic direction,
which can be characterized as the solution of a quadratic optimization problem.
In this respect AdaGrad is similar to GD,
even though their asymptotic directions are usually different.
The difference between them also lies in the stability of their asymptotic directions.
The asymptotic direction of GD is uniquely determined
by the predictors $\bm{x}_n$'s
and independent of initialization and learning rate, as well as rotation of coordinate system,
while the asymptotic direction of AdaGrad is likely to be affected by those factors.

In spite of all these findings,
we still do not know whether the asymptotic direction of AdaGrad will change
for various initialization or different learning rates.
Furthermore,
we hope our approach can be applied to the research on the implicit biases of
other adaptive methods such as AdaDelta, RMSProp, and Adam.

\medskip

\bibliography{bibfile}

\newpage

\begin{center}
  {\Large \bf Appendix}
\end{center}

To simplify notation, we denote
$$
   S_{i}(t) = \sum_{\tau=0}^{t} g_{i}(\tau)^2  \,,
$$
for all $ i \in \lb 1,\cdots, p\rb $ and $ t =0, 1, 2, \cdots \,.$

\bigskip

\noindent
{\em Proof of \bf Lemma 3.1. }
Since $ l $ is $\beta-$smooth, so is $ \mathcal{L} \,.$
Thus we have
\begin{eqnarray*}
  &  &\mathcal{L}( \bm{w}(t+1)) \\
  &\leq & \mathcal{L}( \bm{w}(t)) + \nabla \mathcal{L}( \bm{w}(t))\lp \bm{w}(t+1)-\bm{w}(t)\rp \\
  & & + \frac{\beta}{2}\| \bm{w}(t+1)-\bm{w}(t)  \|^2    \nonumber \\
  &=& \mathcal{L}( \bm{w}(t)) - \eta \bm{g}(t)^T \lp \bm{h}(t)\odot\bm{g}(t) \rp \\
  & & + \frac{\beta\eta^{2}}{2}\| \bm{h}(t)\odot\bm{g}(t) \|^2 \,.
\end{eqnarray*}
Thus
\begin{eqnarray}
  & & \mathcal{L}( \bm{w}(t)) - \mathcal{L} ( \bm{w}(t+1))       \nonumber  \\
 &\geq & \eta \bm{g}(t)^T \lp \bm{h}(t)\odot\bm{g}(t) \rp
  - \frac{\beta\eta^{2}}{2}\| \bm{h}(t)\odot\bm{g}(t) \|^2     \nonumber \\
  &=&  \eta \sum_{i=1}^{p}\frac{g_{i}(t)^2}{\sqrt{S_{i}(t) + \epsilon}}
    -\frac{\beta\eta^2}{2}\sum_{i=1}^{p}\frac{g_{i}(t)^2}{S_{i}(t)+ \epsilon}  \nonumber \\
  &=&  \eta \sum_{i=1}^{p} \lp 1 - \frac{\beta\eta}{2\sqrt{S_{i}(t)+ \epsilon}} \rp  \frac{g_{i}(t)^2}{\sqrt{S_{i}(t)+ \epsilon}} \nonumber  \\
  & > & 0 .   \label{3}
\end{eqnarray}
\eop

\bigskip

\noindent
{\em Proof of \bf Lemma 3.2. }
We use reduction of absurdity.
Suppose
\begin{equation*}
\sum_{t=1}^{\infty}\| \bm{g}(t) \|^2 \ = \ \infty \,.
\end{equation*}
Then there is some $ k\in\lb 1,\cdots, p\rb $ such that
\begin{equation}\label{4}
   \lim_{t\ra\infty} S_{k}(t) = \sum_{t=1}^{\infty}g_{k}(t)^2 = \infty \,.
\end{equation}
Thus we can find a time $\ t_0 $ such that, for all $\ t > t_0 \,,$
$$
    S_{i}(t) > \max(\beta\eta,\ 1) \,.
$$
Noting that positive series
$$ \sum_{t=1}^{\infty} a_t \,,\ \ \
\sum_{t=1}^{\infty} \frac{a_t}{ a_1 + \cdots + a_t + \epsilon}
$$
converge or diverge simultaneously, so we obtain from \eqref{4}
\begin{equation*}
    \sum_{t=0}^{\infty} \frac{g_{k}(t)^2}{S_{k}(t) + \epsilon} = \infty\,.
\end{equation*}
Therefore,
\begin{eqnarray}
   & & \sum_{t=0}^{\infty} \lp 1 - \frac{\beta\eta}{2\sqrt{S_{k}(t) + \epsilon}} \rp  \frac{g_{k}(\tau)^2}{\sqrt{S_{k}(t) + \epsilon}}         \nonumber  \\
   & = & \sum_{t=0}^{t_0} \lp 1 - \frac{\beta\eta}{2\sqrt{S_{k}(t) + \epsilon}} \rp  \frac{g_{k}(\tau)^2}{\sqrt{S_{k}(t) +\epsilon}}         \nonumber  \\
   &  &+ \sum_{t> t_0} \lp 1 - \frac{\beta\eta}{2\sqrt{S_{k}(t) + \epsilon}} \rp  \frac{g_{k}(\tau)^2}{\sqrt{S_{k}(t) +\epsilon}}         \nonumber  \\
   & \geq  &  C + \frac{1}{2} \sum_{t> t_0} \frac{g_{k}(t)^2}{\sqrt{S_{k}(t) + \epsilon}}        \nonumber \\
   & \geq  &  C + \frac{1}{2} \sum_{t> t_0} \frac{g_{k}(t)^2}{S_{k}(t) + \epsilon}  \nonumber \\
   & = & C + \infty = \infty\,,  \label{5}
\end{eqnarray}
where the constant
$$
   C = \sum_{t=0}^{t_0} \lp 1 - \frac{\beta\eta}{2\sqrt{S_{k}(t) + \epsilon}} \rp  \frac{g_{k}(\tau)^2}{\sqrt{S_{k}(t) +\epsilon}} \,.
$$

On the other hand, from \eqref{3} we have
\begin{eqnarray*}
   & & \sum_{\tau=0}^{t} \lp 1 - \frac{\beta\eta}{2\sqrt{S_{k}(t) + \epsilon}} \rp  \frac{g_{k}(\tau)^2}{\sqrt{S_{k}(t) + \epsilon}}  \\
   &\leq & \sum_{\tau=0}^{t} \sum_{i=1}^{p} \lp 1 - \frac{\beta\eta}{2\sqrt{S_{i}(\tau)+ \epsilon}} \rp  \frac{g_{i}(\tau)^2}{\sqrt{S_{i}(\tau)+ \epsilon}}     \\
   &= & \sum_{i=1}^{p}\lb \sum_{\tau=0}^{t} \lp 1 - \frac{\beta\eta}{2\sqrt{S_{i}(\tau)+ \epsilon }} \rp  \frac{g_{i}(\tau)^2}{\sqrt{S_{i}(\tau)+ \epsilon}} \rb   \\
   &\leq & \frac{1}{\eta} \lp \mathcal{L}(\bm{w}(0)) - \mathcal{L}(\bm{w}(t+1)) \rp \leq  \frac{ \mathcal{L}(\bm{w}(0))}{\eta} \,,
\end{eqnarray*}
implying, for sufficiently small $\eta $,
\begin{equation*}
   \sum_{t=0}^{\infty} \lp 1 - \frac{\beta\eta}{2\sqrt{S_{k}(t) + \epsilon}} \rp  \frac{g_{k}(\tau)^2}{\sqrt{S_{k}(t) + \epsilon}}\leq \frac{ \mathcal{L}(\bm{w}(0))}{\eta} ,
\end{equation*}
which contradicts to \eqref{5}.
\eop

\bigskip

\noindent
{\em Proof of \bf Lemma 3.3. }
Lemma 3.2 implies (i), which yields (ii).

To prove (iii), we use reduction to absurdity.
Assume
$$
    \underset{t\ra\infty}{\overline{\lim}} \mathcal{L}(\bm{w}(t)) = c > 0 .
$$
Then there exists an index $ m\in\lb 1,\cdots, N\rb $ such that
$$
   \underset{t\ra\infty}{\overline{\lim}}  l\lp \bm{w}(t)^T \bm{x}_{m} \rp \geq \frac{c}{N} > 0 .
$$
By Assumption 2, we have $ l(u)\ra 0 \ \ (u\ra \infty) $.
Thus we can find a constant $ M > 0 $ such that
$$
   \underset{t\ra\infty}{\underline{\lim}}  \bm{w}(t)^T \bm{x}_{m} \leq  M ,
$$
which implies that there exists a sequence of times
$$
    t_{1} < t_{2} < t_{3} < \cdots
$$
such that
$$
   \underset{k\ra\infty}{\lim}  \bm{w}\lp t_{k} \rp^T \bm{x}_{m}  = \gamma  \leq  M .
$$
Choose a vector $ \bm{w}_{\ast} $ such that $ \bm{w}_{\ast}^T \bm{x}_{n} > 0$ for all $ n\in \lb 1,\cdots, N\rb $.
Noting that $ -l' > 0 $, we have
\begin{eqnarray*}
  - \bm{w}_{\ast}^T \bm{g}(t)  &=& - \sum_{n=1}^{N} l' \lp \bm{w}(t)^T \bm{x}_{n} \rp \bm{w}_{\ast}^T \bm{x}_{n} \\
  &\geq &   l' \lp \bm{w}(t)^T \bm{x}_{m} \rp \bm{w}_{\ast}^T \bm{x}_{m}
\end{eqnarray*}
Thus
\begin{eqnarray}
   &  & \underset{k\ra\infty}{\overline{\lim}} - \bm{w}_{\ast}^T \bm{g}\lp t_{k} \rp   \nonumber \\
   & \geq & \underset{k\ra\infty}{\overline{\lim}}l' \lp \bm{w}\lp t_{k} \rp^T \bm{x}_{m} \rp \bm{w}_{\ast}^T \bm{x}_{m}    \nonumber \\
   & = & \lp \bm{w}_{\ast}^T \bm{x}_{m} \rp \underset{k\ra\infty}{\lim } l' \lp \bm{w}\lp t_{k} \rp^T \bm{x}_{m} \rp     \nonumber \\
   & = & \lp \bm{w}_{\ast}^T \bm{x}_{m} \rp l' \lp \gamma \rp > 0 .   \label{6}
\end{eqnarray}
Note that
$$
 \left\| \bm{g}\lp t \rp \right \| \ra 0\ \ (t\ra \infty)
$$
implies
$$
   - \bm{w}_{\ast}^T \bm{g}\lp t_{k} \rp \leq   \left\| \bm{w}_{\ast}\right\|  \left\| \bm{g}\lp t_{k} \rp \right \| \ra 0\ \ (k\ra \infty) ,
$$
which contradicts to \eqref{6}, meaning (iii) has to be true.

(iv) follows from (iii).
(v) follows directly from (iv).
\eop

\bigskip

\noindent
{\em Proof of \bf Theorem 3.1. }
By Lemma 3.2
$ \lb h_{i}(t) \rb_{t=0}^{\infty} $
is decreasing and has a lower bound
$$
    \frac{1}{\sqrt{S+ \epsilon}} > 0 \,,
$$
where
$$
   S = \lim_{t\ra\infty}S_{i}(t) \leq \sum_{t=0}^{\infty}\| \bm{g}(t) \|^2 < \infty  \,,
$$
then converges,
for each $ i\in\lb 1,\cdots, p\rb \,.$
\eop

\bigskip

\noindent
{\bf Lemma A.1.}\hspace{4pt}
Let  $ \ \bm{a}\,,\ \bm{b} = (b_1, \cdots, b_p)\in \Rp \,.$
Then the following relations hold.

(i) Associativity. $  \lp \bm{a}\odot \bm{b}\rp \odot \bm{v} = \bm{a}\odot \lp \bm{b}\odot \bm{v} \rp   \,.$

(ii) Commutativity. $  \bm{a}\odot \bm{b} = \bm{b}\odot \bm{a} \,;$

(iii) Distributivity. $  \bm{a}\odot \lp \bm{b} + \bm{c} \rp = \bm{a}\odot\bm{b} + \bm{a} \odot \bm{c}    \,.$

(iv)
$\
 \underset{i}{\min}\ |b_{i}|\ \|\bm{a}\| \leq \| \bm{b}\odot\bm{a} \| \leq \underset{i}{\max}\ |b_{i}| \ \|\bm{a}\|\ \,.
$

\medskip

\noindent
{\em Proof. }
Obviously.
\eop

\bigskip

\noindent
{\em Proof of \bf Lemma 3.4. }
It directly follows from Lemma 3.3.
\eop

\bigskip

\noindent
{\bf Lemma A.2.}\hspace{4pt}
For $\ t = 0,1,2\cdots\,,$
\begin{equation*}
        \bm{\delta}(t)^T\widehat{\bm{u}} = \| P \bm{\delta}(t)\| \geq \frac{\|\bm{\delta}(t) \|}{\underset{n}{\max}\ \|\bm{\xi}_{n}\|}\,,
\end{equation*}
where
\begin{equation*}
    \widehat{\bm{u}} = \underset{\bm{u}^T \bm{\xi}_{n}\geq 1,\, \forall n}{\arg\min} \| \bm{u} \|^{2} .
\end{equation*}

\medskip

\noindent
{\em Proof. }
From Assumption 2. we have
$\
 -  l' \lp \bm{v}^{T} \bm{\xi}_{n}\rp > 0 \,.
$
By the definition of $ \widehat{\bm{u}} $ we have
$$
   \bm{\xi}_{n}^{T}\widehat{\bm{u}} \geq 1\ \ \ (n = 1, \cdots, N).
$$
Thus
\begin{equation*}
  \bm{\delta}(t)^{T}\widehat{\bm{u}} = - \eta \nabla \lambda \lp \bm{v}(t) \rp^{T}\widehat{\bm{u}}
  = -\sum_{n=1}^{N} l' \lp \bm{v}^{T} \bm{\xi}_{n}\rp\bm{\xi}_{n}^{T}\widehat{\bm{u}} \geq -\sum_{n=1}^{N} l' \lp \bm{v}^{T} \bm{\xi}_{n}\rp > 0 .
\end{equation*}
Note that $ l' < 0 $.
We have
\begin{eqnarray*}
   \| \bm{\delta}(t) \|
   &=& \left\|\eta \sum_{n=1}^{N}l'\lp \bm{v}(t)^T \bm{\xi}_{n}\rp \bm{\xi}_{n} \right\| \leq -\eta\sum_{n=1}^{N}l'\lp \bm{v}(t)^T \bm{\xi}_{n}\rp \left\| \bm{\xi}_{n} \right\| \\
   &\leq &\ \underset{n}{\max}\ \|\bm{\xi}_{n}\|\lp -\eta\sum_{n=1}^{N} l'\lp \bm{v}(t)^T \bm{\xi}_{n}\rp \rp \,,
\end{eqnarray*}
or
\begin{equation}\label{12}
   -\eta\sum_{n=1}^{N} l'\lp \bm{v}(t)^T \bm{\xi}_{n}\rp \geq \frac{\| \bm{\delta}(t) \|}{\underset{n}{\max}\ \|\bm{\xi}_{n}\|}  .
\end{equation}
On the other hand,
\begin{eqnarray*}
   P \bm{\delta}(t)
   &=& -\eta P \sum_{n=1}^{N}l'\lp \bm{v}(t)^T \bm{\xi}_{n}\rp \bm{\xi}_{n} = -\eta\sum_{n=1}^{N}l'\lp \bm{v}(t)^T \bm{\xi}_{n}\rp P\bm{\xi}_{n} \\
   &=& -\eta \sum_{n=1}^{N}l'\lp \bm{v}(t)^T \bm{\xi}_{n}\rp \lp \bm{\xi}_{n}^{T}\widehat{\bm{u}} \rp \widehat{\bm{u}} .
\end{eqnarray*}
Noting $ \bm{\xi}_{n}^{T}\widehat{\bm{u}}\geq 1\ (n\in\lb 1,\cdots, N\rb) ,$ from \eqref{12} we obtain
\begin{eqnarray*}
   \| P \bm{\delta}(t) \|
   &=& -\eta\sum_{n=1}^{N}l'\lp \bm{v}(t)^T \bm{\xi}_{n}\rp \lp \bm{\xi}_{n}^{T}\widehat{\bm{u}} \rp
   \geq -\eta\sum_{n=1}^{N}l'\lp \bm{v}(t)^T \bm{\xi}_{n}\rp \geq \frac{\| \bm{\delta}(t) \|}{\underset{n}{\max}\ \|\bm{\xi}_{n}\| } .
\end{eqnarray*}
\eop

\bigskip

\noindent
{\bf Lemma A.3.}\hspace{4pt}
For sufficiently large $\ t \,,$
\begin{equation}\label{13}
   \frac{1}{2}\| \bm{\delta}(t) \| \leq \| \bm{d}(t) \| \leq  \frac{3}{2}\| \bm{\delta}(t) \|  ,
\end{equation}
\begin{equation}\label{14aaa}
    \| P \bm{d}(t)\| \geq  \frac{ \|\bm{d}(t) \|}{4\,\underset{n}{\max}\ \|\bm{\xi}_{n}\|} \,,
\end{equation}
\begin{equation}\label{15}
    \bm{d}(t)^{T}\widehat{\bm{u}} = \| P\bm{d}(t) \| > 0 \,.
\end{equation}

\medskip

\noindent
{\em Proof. }
Let $ \bm{\beta}(t) = \lp \beta_{1}(t),\cdots, \beta_{p}(t) \rp^T .$
Noting that
$$
 \|\bm{\beta}(t) -\bm{1}\| \ra 0 \ \ (t \ra \infty) ,
$$
we can find some $ t_0 $ such that for $ t \geq t_0 \,,$
\begin{equation}\label{16}
  \frac{1}{2} \leq \underset{i}{\min}\ |\beta_{i}(t)| \leq \underset{i}{\max}\ |\beta_{i}(t)| \leq \frac{3}{2}
\end{equation}
and
\begin{equation}\label{17}
    \underset{i}{\max}\ | \beta_{i}(t) - 1 | < \frac{1}{2\ \underset{n}{\max}\|\bm{\xi}_{n}\|} \,.
\end{equation}
The inequality \eqref{13} follows directly from \eqref{16}.
On the other hand,
\begin{equation*}
   P \bm{d}(t)
   = P \lp \bm{\beta}(t)\odot \bm{\delta}(t) \rp
   = P \bm{\delta}(t) + P \lp \lp \bm{\beta}(t) -\bm{1} \rp \odot \bm{\delta}(t) \rp
\end{equation*}
By \eqref{17} we have
\begin{equation*}
     \|P \lp \lp \bm{\beta}(t) -\bm{1} \rp \odot \bm{\delta}(t) \rp\|
   \leq \left\|\lp \bm{\beta}(t) -\bm{1} \rp \odot \bm{\delta}(t) \right\|
   \leq \underset{i}{\max}\ | \beta_{i}(t) - 1 | \left\| \bm{\delta}(t) \right\|  \leq \frac{\left\|\bm{\delta}(t)\right\|}{2\ \underset{n}{\max}\|\bm{\xi}_{n}\|} .
\end{equation*}
Hence
\begin{eqnarray*}
   \| P\bm{d}(t)\|
   &=& \left\| P \bm{\delta}(t) + P \lp \lp \bm{\beta}(t) -\bm{1} \rp \odot \bm{\delta}(t) \rp\right\| \\
   &\geq & \left\| P \bm{\delta}(t) \right\| - \left\|  P \lp \lp \bm{\beta}(t) -\bm{1} \rp \odot \bm{\delta}(t) \rp \right\|   \\
   &\geq & \frac{\|\bm{\delta}(t) \|}{\underset{n}{\max}\ \|\bm{\xi}_{n}\|} -
   \frac{\left\|\bm{\delta}(t)\right\|}{2\ \underset{n}{\max}\|\bm{\xi}_{n}\|} = \frac{\|\bm{\delta}(t) \| }{2\underset{n}{\max}\ \|\bm{\xi}_{n}\|} .
\end{eqnarray*}
Thus \eqref{14aaa} follows from the left part of \eqref{13}.

Noting that
\begin{align*}
  \bm{d}(t)^{T}\widehat{\bm{u}}
   & = \bm{\delta}(t)^{T}\widehat{\bm{u}} + \lp  \bm{\beta}(t)- \bm{1} \rp\odot \bm{\delta}(t)^{T}\widehat{\bm{u}} \\
   & \geq \bm{\delta}(t)^{T}\widehat{\bm{u}} -\left|\lp \bm{\beta}(t)- \bm{1} \rp\odot \bm{\delta}(t)^{T}\widehat{\bm{u}} \right|      \\
   & = \left\|P \bm{\delta}(t)\right\|  - \left\|  P \lp \lp \bm{\beta}(t) -\bm{1} \rp \odot \bm{\delta}(t) \rp \right\|  \\
   & \geq \frac{\|\bm{\delta}(t) \|}{\underset{n}{\max}\ \|\bm{\xi}_{n}\|} - \frac{\|\bm{\delta}(t) \| }{2\underset{n}{\max}\ \|\bm{\xi}_{n}\|}    \\
   & = \frac{\|\bm{\delta}(t) \| }{2\underset{n}{\max}\ \|\bm{\xi}_{n}\|} > 0 \,,
\end{align*}
we obtain \eqref{15}.
\eop

\bigskip

\noindent
{\bf Lemma A.4.}\hspace{4pt}
For sufficiently large $\ t \,,$
$$
    \| P \bm{v}(t)\| \geq \frac{\|\bm{v}(t) \|}{8\underset{n}{\max}\ \|\bm{\xi}_{n}\|} \,.
$$

\medskip

\noindent
{\em Proof. }
By Lemma A.3 there there exists some $\ t_0 $ such that for $ t \geq t_0 $,
\begin{equation*}
    \| P \bm{d}(t)\| \geq  \frac{\|\bm{d}(t) \|}{4\underset{n}{\max}\ \|\bm{\xi}_{n}\|} \,.
\end{equation*}
Note that $\ \|\bm{v}(t)\| \ra \infty \,,$
which implies
$$
   \|\bm{d}(t_{0})\| + \cdots + \|\bm{d}(t)\| \geq \|\bm{v}(t)\| - \|\bm{v}(t_0)\| \ra \infty \ \ (t\ra \infty)\,.
$$
Thus there exists some $\ t_1 > t_0 $ such that for $ t > t_1 $,
$$
   \|\bm{d}(t_{0})\| + \cdots + \|\bm{d}(t)\| > 2 \| \bm{v}(t_{0}) \| \,,
$$
Hence, meanwhile,
\begin{eqnarray*}
   \| P \bm{v}(t)\|
   &=& \| P \bm{v}(t_{0}) + P \bm{d}(t_{0}) + \cdots + P\bm{d}(t-1) \| \\
   &= & \| P \bm{v}(t_{0})\| + \|P \bm{d}(t_{0})\| + \cdots + \|P\bm{d}(t-1) \| \\
   &\geq & \frac{1}{4\underset{n}{\max}\ \|\bm{\xi}_{n}\|}\lp \|\bm{d}(t_{0})\| + \cdots + \|\bm{d}(t-1) \| \rp \\
   &\geq & \frac{1}{8\underset{n}{\max}\ \|\bm{\xi}_{n}\|}\lp \| \bm{v}(t_{0}) \| + \|\bm{d}(t_{0})\| + \cdots + \|\bm{d}(t-1) \| \rp \\
   &\geq & \frac{1}{8\underset{n}{\max}\ \|\bm{\xi}_{n}\|} \| \bm{v}(t_{0}) + \bm{d}(t_{0}) + \cdots + \bm{d}(t-1) \| \\
   &  =  & \frac{\| \bm{v}(t)\|}{8\underset{n}{\max}\ \|\bm{\xi}_{n}\|} .
\end{eqnarray*}
\eop

\bigskip

\noindent
{\bf Lemma A.5.}\hspace{4pt}
Let
$$
   \mathcal{K} = \lb n:\ \bm{\xi}_{n}^{T}\widehat{\bm{u}} = 1 \rb .
$$
Then there is a set of nonnegative numbers $ \lb \alpha_{n}:\ n\in \mathcal{K} \rb $
such that
$$
    \widehat{\bm{u}} = \sum_{n\in\mathcal{K}} \alpha_{n}\bm{\xi}_{n} \,.
$$

\medskip

\noindent
{\em Proof. }
This is Lemma 12 in Appendix B of Soudry et al., [2018].
\eop

\bigskip

\noindent
{\em Proof of \bf Lemma 3.5. }
Since for each $ n\in \lb 1,\cdots, N\rb $,
$$
 \bm{v}(t)^T \bm{\xi}_{n}\ra\infty \ \ (t\ra \infty) ,
$$
we have, for sufficiently large $ t $,
\begin{eqnarray}
   - l'\lp \bm{v}(t)^T \bm{\xi}_{n} \rp
   &=& c e^{- a \bm{v}(t)^T \bm{\xi}_{n}} - r \lp \bm{v}(t)^T \bm{\xi}_{n} \rp  \nonumber\\
   &\geq & c e^{- a \bm{v}(t)^T \bm{\xi}_{n}} - e^{- (a+b) \bm{v}(t)^T \bm{\xi}_{n}}  \nonumber\\
   & = & e^{- a \bm{v}(t)^T \bm{\xi}_{n}} \lp c - e^{- b \bm{v}(t)^T \bm{\xi}_{n}}\rp  \nonumber\\
   & = & \frac{c}{2} e^{- a \bm{v}(t)^T \bm{\xi}_{n}}.  \label{18}
\end{eqnarray}
Similarly, we can prove for sufficiently large $ t $,
\begin{equation}\label{19}
   - l'\lp \bm{v}(t)^T \bm{\xi}_{n} \rp \leq 2c e^{- a \bm{v}(t)^T \bm{\xi}_{n}}.
\end{equation}

Denote
$$
   p(t) = \bm{v}(t)^{T}\widehat{\bm{u}} ,\ \ \ \bm{q}(t) = Q\bm{v}(t) .
$$
Thus we have
\begin{equation*}
    \bm{v}(t) = p(t)\widehat{\bm{u}} + \bm{q}(t) .
\end{equation*}
Denote
$$
   u_{n} = \widehat{\bm{u}}^T \bm{\xi}_{n} ,\ \ \  q_{n,t} = \bm{q}(t)^T \bm{\xi}_{n} .
$$
We then have
\begin{eqnarray*}
   & & \bm{q}(t)^{T}\bm{\delta}(t) \\
   &=& -\eta \bm{q}(t)^{T} \sum_{n=1}^{N} l'\lp \bm{v}(t)^T \bm{\xi}_{n} \rp \bm{\xi}_{n}   \\
   &=&  -\eta \sum_{n=1}^{N}l' \lp \bm{v}(t)^T \bm{\xi}_{n} \rp \bm{q}(t)^{T}\bm{\xi}_{n}    \\
   &=&  -\eta \sum_{n=1}^{N}l' \lp \bm{v}(t)^T \bm{\xi}_{n} \rp  q_{n,t}     \\
   &\leq &  -\eta \sum_{n:\ q_{n,t} > 0} l' \lp \bm{v}(t)^T \bm{\xi}_{n} \rp  q_{n,t} .
\end{eqnarray*}
Applying \eqref{19} we obtain
\begin{eqnarray*}
   \bm{q}(t)^{T}\bm{\delta}(t)
   &\leq&  \eta \sum_{n:\ q_{n,t} > 0}  2 c e^{ - a\bm{v}(t)^T \bm{\xi}_{n} } q_{n,t}     \\
   &=&   2c \eta \sum_{n:\ q_{n,t} > 0}  e^{ - a\lp p(t)\widehat{\bm{u}} + \bm{q}(t) \rp^T \bm{\xi}_{n} } q_{n,t}  \\
   &=& 2c\eta \sum_{n:\ q_{n,t} > 0} e^{ - a p(t)u_{n}} e^{-a q_{n,t}  } q_{n,t}   \\
  &\leq &  \frac{2c\eta }{a e} \sum_{n:\ q_{n,t} > 0} e^{ - a p(t)u_{n}}  \\
  &\leq &  \frac{2c\eta N}{a e} e^{ - a p(t)} .
\end{eqnarray*}
The last step is derived from $ u_{n}\geq 1 $ for $ n\in\lb 1,\cdots, N\rb $.

On the other hand,
by Lemma A.5 there is a set of positive coefficients
$\lb \alpha_{n}: n\in \mathcal{K} \rb $,
where
$\
   \mathcal{K} = \lb n:\ u_{n} = 1 \rb  ,
$
such that
$$
   \widehat{\bm{u}} = \sum_{n\in \mathcal{K}} \alpha_{n}\bm{\xi}_{n} \,.
$$
Thus
$$
   0 = \bm{q}(t)^{T}\widehat{\bm{u}} = \sum_{n\in \mathcal{K}} \alpha_{n}\bm{q}(t)^{T}\bm{\xi}_{n} \,,
$$
implying there is at least one index $\ k\in \mathcal{K} $ such that
$$
  q_{t,k} = \bm{q}(t)^{T}\bm{\xi}_{k} \leq 0 \,.
$$
Hence
\begin{eqnarray*}
  \|P \bm{\delta}(t)\|
   &=& \eta \left\| P \sum_{n=1}^{N} l'\lp \bm{v}(t)^T \bm{\xi}_{n} \rp \bm{\xi}_{n}   \right\| \\
   &=& \eta \left\|  \sum_{n=1}^{N} l'\lp \bm{v}(t)^T \bm{\xi}_{n} \rp P\bm{\xi}_{n}  \right\|  \\
   &=& \eta \left\| \sum_{n=1}^{N} l'\lp \bm{v}(t)^T \bm{\xi}_{n} \rp u_{n}\widehat{\bm{u}} \right\| \\
   &=& -\eta \sum_{n=1}^{N} u_{n}  l'\lp \bm{v}(t)^T \bm{\xi}_{n} \rp  \\
   &\geq & -\eta \sum_{n=1}^{N} l'\lp \bm{v}(t)^T \bm{\xi}_{n} \rp    \\
   & > &  -\eta l'\lp \bm{v}(t)^T \bm{\xi}_{k} \rp    \,.
\end{eqnarray*}
Noting that $ u_{k} \geq 1 $, $ q_{t,k}\leq 0 $,
and the estimation \eqref{18},
we obtain
\begin{eqnarray*}
   \|P \bm{\delta}(t)\| &>& -\eta l'\lp \bm{v}(t)^T \bm{\xi}_{k} \rp \\
   &\geq & \frac{c\eta}{2} e^{- a \bm{v}(t)^T \bm{\xi}_{k}} \\
   & = & \frac{c\eta}{2} e^{- a p(t) u_{k}} e^{-a q_{t,k}}  \\
   &\geq & \frac{c\eta}{2} e^{- a p(t)} .
\end{eqnarray*}
Thus
\begin{eqnarray*}
   \bm{q}(t)^{T} \bm{\delta}(t)
   &\leq & \eta \lp c  + 1 \rp \frac{N}{a e} e^{ - a p(t)} \\
   &\leq & \frac{2 N\lp c  + 1 \rp}{a c e} \|P \bm{\delta}(t)\| \\
   &\leq & \frac{2 N\lp c  + 1 \rp }{a c e} \|\bm{\delta}(t)\| \\
   &< & \varepsilon \| \bm{q}(t) \|\|\bm{\delta}(t)\|\,,
\end{eqnarray*}
for
$\  \| \bm{q}(t) \| > 2(ace\varepsilon)^{-1}N(c+1) \,. $
\eop

\bigskip

\noindent
{\em Proof of \bf Lemma 3.6. }
Again we denote $\ \bm{q}(t) = Q\bm{v}(t).$
By Lemma 3.5 we can choose a number $ R > 0 $ such that
for sufficiently large $\,t $ and $\ \| \bm{q}(t)\| \geq R $,
\begin{equation}\label{20}
  \bm{q}(t) ^{T}\bm{\delta}(t) < \frac{\varepsilon}{16} \| \bm{q}(t)\|\| \bm{\delta}(t)\|
\end{equation}
and
\begin{equation}\label{13aaa}
   \frac{1}{2}\| \bm{\delta}(t) \| \leq \| \bm{d}(t) \|
\end{equation}
from \eqref{13}. Noting
$$
 \|\bm{q}(t+1) \|^2 - \|\bm{q}(t) \|^2 = 2 \bm{q}(t)^{T}Q \bm{d}(t) + \| Q d(t) \|^2 ,
$$
we have
\begin{eqnarray*}
 \| \bm{q}(t+1)\| - \| \bm{q}(t)\|
  &=& \frac{\|\bm{q}(t+1) \|^2 - \|\bm{q}(t) \|^2 }{\| \bm{q}(t+1)\| + \| \bm{q}(t)\|}   \\
  &=& \frac{2 \bm{q}(t)^{T}Q \bm{d}(t) + \| Q d(t) \|^2}{\| \bm{q}(t+1)\| + \| \bm{q}(t)\|}  \\
  &\leq & \frac{2 \bm{q}(t)^{T}\bm{d}(t) + \| Q d(t) \|^2}{\| \bm{q}(t)\|} \\
  &= & \frac{2 \bm{q}(t)^{T}\lp \bm{\delta}(t) + ( \bm{\beta}(t) -\bm{1}) \odot  \bm{\delta}(t) \rp + \| Q d(t) \|^2}{\| \bm{q}(t)\|} \\
  &=& \frac{2 \bm{q}(t)^{T}\bm{\delta}(t)}{\| \bm{q}(t)\|}+\frac{2 \bm{q}(t)^{T} ( \bm{\beta}(t) -\bm{1})\odot  \bm{\delta}(t) }{\| \bm{q}(t)\|} + \frac{ \| Q d(t) \|^2}{\| \bm{q}(t)\|} \\
  &\leq & \frac{\varepsilon \|\bm{\delta}(t)\|}{8}+ 2\|( \bm{\beta}(t) -\bm{1}) \odot  \bm{\delta}(t)\| + \frac{ \| d(t) \|^2}{R}
\end{eqnarray*}
Since
$$
   \bm{\beta}(t) = \lp \beta_{1}(t),\cdots, \beta_{p}(t) \rp^T  \ra \bm{1}\ \ \ (t\ra \infty)
$$
and
$$
   \|\bm{d}(t)\| \ra 0\ \ \ (t\ra \infty),
$$
we can see that for sufficiently large $\,t $,
$$
    \underset{i}{\max}\ | \beta_{i}(t) - 1 | < \frac{\varepsilon}{8}
$$
and
$$
    \|\bm{d}(t)\| < \frac{R\varepsilon}{8} .
$$
Now we have
\begin{align*}
  \|( \bm{\beta}(t) -\bm{1}) \odot  \bm{\delta}(t)\| &\leq \underset{i}{\max}\ | \beta_{i}(t) - 1 |\| \bm{\delta}(t)\| \leq \frac{\varepsilon}{8} \| \bm{\delta}(t)\| \,,  \\
   & \frac{ \| d(t) \|^2}{R} \leq \frac{\varepsilon}{8} \| \bm{\delta}(t)\|  .
\end{align*}
By \eqref{13aaa}, we obtain
\begin{eqnarray*}
  & & \| \bm{q}(t+1)\| - \| \bm{q}(t)\|  \\
  &\leq & \frac{\varepsilon \|\bm{\delta}(t)\|}{8}+ 2\|( \bm{\beta}(t) -\bm{1}) \odot  \bm{\delta}(t)\| + \frac{ \| d(t) \|^2}{R} \\
  &\leq & \frac{\varepsilon \|\bm{\delta}(t)\|}{8}+  \frac{\varepsilon \|\bm{\delta}(t)\|}{4} +  \frac{\varepsilon \|\bm{\delta}(t)\|}{8} \\
  &=&  \frac{\varepsilon \|\bm{\delta}(t)\|}{2} \leq \varepsilon \|\bm{d}(t)\| .
\end{eqnarray*}
\eop

\bigskip

\noindent
{\em Proof of \bf Lemma 3.7. }
Since $ \| \bm{d}(\tau) \| \ra 0 \ \ (\tau\ra\infty) $, we we can find a time $ t_{0} $ such that for $ \tau \geq t_{0} $,
$$
    \|\bm{d}(\tau)\| \leq 1 .
$$
By Lemma A.3 we can find a time $ t_{1} \geq t_{0} $ such that for $ \tau \geq t_{1} $,
\begin{equation}\label{21}
 \|\bm{d}(\tau)\| \leq \lp 4\,\underset{n}{\max}\ \|\bm{\xi}_{n}\|\rp \|P\bm{d}(\tau)\| .
\end{equation}
Given $ \varepsilon > 0\,,$ by Lemma 3.6
we can choose $ R \geq 1 $ and $ t_{2} \geq t_{1} $ such that for $ \tau \geq t_{2} $ and  $\ \|\bm{q}(\tau)\| \geq R $,
$$
   \| \bm{q}(\tau+1)\| - \| \bm{q}(\tau)\| \leq \varepsilon \|\bm{d}(\tau)\|  .
$$
Since $ \| \bm{v}(\tau)\|\ra \infty \ \ (\tau\ra\infty) $, we can choose
$ t_{3} \geq t_{2} $ such that for $ \tau \geq t_{3} $,
\begin{equation}\label{22}
    \| \bm{v}(\tau)\|^{-1} R < \varepsilon
\end{equation}
and
\begin{equation}\label{23}
   \| \bm{v}(\tau)\|^{-1}\lp \|\bm{q}(t_{2})\| + 4\,\varepsilon\,\underset{n}{\max}\ \|\bm{\xi}_{n}\| \|P\bm{v}(t_{2}) \|\rp < \varepsilon .
\end{equation}

Now let $ t \geq t_{3} $.
To simplify notation we denote
$$
    \xi^{\ast} = \underset{n}{\max}\ \|\bm{\xi}_{n}\| .
$$

\textbf{Case 1.}
If $ \|\bm{q}(t)\| < R , $
then from \eqref{22} we directly obtain
\begin{equation}\label{24}
   \| \bm{v}(t)\|^{-1} \|\bm{q}(t)\| <\varepsilon.
\end{equation}

\textbf{Case 2.}
If for each $ \tau \in \lb t_{2},\cdots, t \rb $,
$\ \|\bm{q}(\tau)\| \geq R ,$
then from \eqref{21},
\begin{eqnarray*}
    \|\bm{q}(t)\|
    &= &   \|\bm{q}(t_{2})\| + \sum_{\tau = t_{2}}^{t-1} \lp \|\bm{q}(\tau + 1)\| - \| \bm{q}(\tau)\| \rp  \\
    &\leq & \|\bm{q}(t_{2})\| + \varepsilon \lp \|\bm{d}(t_{2})\|+\cdots+\|\bm{d}(t-1)\| \rp \\
    &\leq & \|\bm{q}(t_{2})\| + 4\,\varepsilon\,\xi^{\ast} \lp \|P\bm{d}(t_{2})\|+\cdots+\|P\bm{d}(t-1)\| \rp   \\
    & =   & \|\bm{q}(t_{2})\| + 4\,\varepsilon\,\xi^{\ast} \lp \|P\bm{d}(t_{2})+\cdots+ P\bm{d}(t-1)\| \rp   \\
    & =   & \|\bm{q}(t_{2})\| + 4\,\varepsilon\,\xi^{\ast} \left\|P\lp \bm{d}(t_{2})+\cdots+ \bm{d}(t-1)\rp\right\|   \\
    & =   & \|\bm{q}(t_{2})\| + 4\,\varepsilon\,\xi^{\ast} \left\|P\bm{v}(t)- P\bm{v}(t_{2})\right\|   \\
    &\leq & \|\bm{q}(t_{2})\| + 4\,\varepsilon\,\xi^{\ast} \lp \|P\bm{v}(t_{2}) \| +\| P\bm{v}(t)\| \rp \,.
\end{eqnarray*}
From \eqref{23} we have
\begin{eqnarray}
   \| \bm{v}(t)\|^{-1} \|\bm{q}(t)\| \nonumber
   &\leq & \| \bm{v}(t)\|^{-1}\lp \|\bm{q}(t_{2})\| + 4\,\varepsilon\,\xi^{\ast} \|P\bm{v}(t_{2}) \|\rp +  4\,\varepsilon\,\xi^{\ast} \| \bm{v}(t)\|^{-1} \| P\bm{v}(t)\| \\
   &< & \varepsilon + 4\,\varepsilon\,\xi^{\ast} = \lp 1 + 4\,\xi^{\ast} \rp \varepsilon . \label{25}
\end{eqnarray}

\textbf{Case 3.}
If $ \|\bm{q}(t)\| \geq R $ and
there is a time $ t_{\ast}\in \lb t_{2},\cdots, t-1 \rb $ such that
$$
 \|\bm{q}(t_{\ast})\| < R ,
$$
then we can find the time $ t^{\ast}\in \lb  t_{\ast},\cdots, t-1 \rb $ such that
$$
 \|\bm{q}(t^{\ast})\| < R
$$
and for each $ \tau \in \lb t^{\ast}+1,\cdots, t \rb $,
$$
 \|\bm{q}(\tau)\| \geq R .
$$
Thus we have
\begin{eqnarray*}
    \|\bm{q}(t)\|
    &= &   \|\bm{q}(t^{\ast})\| + \lp \|\bm{q}(t^{\ast}+1)\|-  \|\bm{q}(t^{\ast})\| \rp
     + \sum_{\tau = t^{\ast}+1}^{t-1} \lp \|\bm{q}(\tau + 1)\| - \| \bm{q}(\tau)\| \rp  \\
    & < & R + \|Q\bm{d}(t^{\ast})\| + \varepsilon \lp \|\bm{d}(t^{\ast}+1)\|+\cdots+\|\bm{d}(t-1)\| \rp \\
    &\leq & R + \|\bm{d}(t^{\ast})\| + 4\,\varepsilon\,\xi^{\ast} \lp \|P\bm{d}(t^{\ast} + 1)\|+\cdots+\|P\bm{d}(t-1)\| \rp   \\
    &\leq & R + \|\bm{d}(t^{\ast})\|+ 4\,\varepsilon\,\xi^{\ast} \lp \|P\bm{d}(t_{2})\|+\cdots+\|P\bm{d}(t-1)\| \rp   \\
    & =   & R + 1 + 4\,\varepsilon\,\xi^{\ast}\lp \|P\bm{d}(t_{2})+\cdots+ P\bm{d}(t-1)\| \rp   \\
    & =   & 2 R + 4\,\varepsilon\,\xi^{\ast} \left\|P\bm{v}(t)- P\bm{v}(t_{2})\right\|   \\
    &\leq & 2 R + 4\,\varepsilon\,\xi^{\ast} \lp \|P\bm{v}(t_{2}) \| +\| P\bm{v}(t)\| \rp \,.
\end{eqnarray*}
Noting \eqref{22} and \eqref{23}, we obtain
\begin{eqnarray}
   \| \bm{v}(t)\|^{-1} \|\bm{q}(t)\| \nonumber
   &\leq & \| \bm{v}(t)\|^{-1}\lp 2R + 4\,\varepsilon\,\xi^{\ast} \|P\bm{v}(t_{2}) \|\rp+  4\,\varepsilon\,\xi^{\ast} \|\bm{v}(t)\|^{-1}\| P\bm{v}(t)\|   \nonumber  \\
   &< & 3\varepsilon + 4\,\varepsilon\,\xi^{\ast} = \lp 3 + 4\,\xi^{\ast} \rp \varepsilon . \label{26}
\end{eqnarray}

Verifying \eqref{24}, \eqref{25} and \eqref{26}, we can see that, in any case,  \eqref{26} is valid.
Since $ \varepsilon $ can be any positive number, we have
$$
   \limt \frac{\|\bm{q}(t)\|}{\| \bm{v}(t)\|} = 0 .
$$
Thus
$$
   \limt \frac{\|P\bm{v}(t)\|}{\| \bm{v}(t)\|} = 1 ,\ \ \
   \limt \frac{\bm{q}(t)}{\| \bm{v}(t)\|} = \bm{0} .
$$
Therefore,
\begin{equation*}
  \limt \frac{\bm{v}(t)}{\| \bm{v}(t)\|}
  = \limt \frac{P\bm{v}(t) + \bm{q}(t) }{\| \bm{v}(t)\|}
  = \limt \frac{P\bm{v}(t)}{\| \bm{v}(t)\|} = \limt \frac{\|P\bm{v}(t)\| }{\| \bm{v}(t)\|} \widehat{\bm{u}}  = \widehat{\bm{u}} \,.
\end{equation*}
\eop

\bigskip

\noindent
{\em Proof of \bf Theorem 3.2. }
By hypothesis
\begin{eqnarray*}
\left\|\bm{h}_{\infty}^{1/2}\odot\bm{w}_{\infty}\right\|^2
  &=&  \underset{\bm{w}^{T}\bm{x}_{n}\geq 1,\, \forall n }{ \min } \left\|\bm{h}_{\infty}^{1/2}\odot\bm{w}\right\|^2 =  \underset{\lp \bm{h}_{\infty}^{1/2}\odot\bm{u}\rp^{T}\bm{x}_{n}\geq 1,\, \forall n }{ \min } \left\|\bm{u}\right\|^2 \\
  &=&  \underset{\bm{u}^{T}\lp \bm{h}_{\infty}^{1/2}\odot\bm{x}_{n}\rp\geq 1,\, \forall n }{ \min } \left\|\bm{u}\right\|^2 =  \underset{\bm{u}^{T}\bm{\xi}_{n} \geq 1,\, \forall n }{ \min } \left\|\bm{u}\right\|^2 \,.
\end{eqnarray*}
Noting that both
$$
    \widehat{\bm{u}} = \underset{\bm{u}^T \bm{\xi}_{n}\geq 1,\, \forall n}{\arg\min} \| \bm{u} \|^{2}  \,.
$$
and $ \bm{w}_{\infty} $ are unique, we must have
$\ \widehat{\bm{u}} = \bm{h}_{\infty}^{-1/2}\odot\bm{w}_{\infty} \,,$
or
$$
    \bm{w}_{\infty} = \bm{h}_{\infty}^{1/2}\odot\widehat{\bm{u}} \,.
$$
From Lemma 3.7 and the relation
\[
      \bm{v}(t) = \bm{h}_{\infty}^{-1/2}\odot \bm{w}(t)\ \ (t = 0,1,2,\cdots) \,,
\]
we obtain
\begin{equation*}
    \bm{w}_{\infty} =  \bm{h}_{\infty}^{1/2}\odot \limt \frac{ \bm{v}(t)}{\| \bm{v}(t)\|} = \limt \frac{\bm{h}_{\infty}^{1/2}\odot \bm{v}(t)}{\| \bm{v}(t)\|} =  \limt \frac{\bm{w}(t)}{\| \bm{v}(t)\|} .
\end{equation*}
Thus
$$
   \limt \frac{\bm{w}(t)}{\| \bm{w}(t)\|} = \limt  \frac{\| \bm{v}(t)\|}{\| \bm{w}(t)\|} \cdot \limt \frac{\bm{w}(t)}{\| \bm{v}(t)\|}  =  \frac{\bm{w}_{\infty}}{\|\bm{w}_{\infty}\|} .
$$
\eop

\bigskip

\noindent
{\bf Lemma A.6.}\hspace{4pt}
Suppose $ N \geq p $ and the $\ p\times N-$matrix
$\ X = \left[ \bm{x}_{1}, \cdots, \bm{x}_{N} \right]\,,$
where
\begin{equation*}\label{2401aaa}
    \bm{x}_{n} = \lp x_{n,1},\cdots, x_{n,p} \rp^T \ \ \ (n = 1,\cdots, N)\,,
\end{equation*}
satisfies the following conditions:

(i) For $\ n = 1,\cdots, p \,,$
\begin{equation*}\label{2402aaa}
   x_{n,i}
   \left\{\begin{array}{ll}
            >0 , & \text{for}\ i=n ,  \\
            <0 , & \text{for}\ i\not=n.
          \end{array}
   \right.
\end{equation*}

(ii) The $\ p\times p-$matrix
$\ X_{p} = \left[ \bm{x}_{1}, \cdots, \bm{x}_{p} \right]\ $ is nonsingular.

(iii) The unique solution $\ \bm{a} = \lp a_{1},\cdots, a_{p} \rp^T\,$
of the linear system in $\ \bm{w} $
\begin{equation}\label{2405aaa}
   \bm{x}_{n}^{T}\bm{w} = 1 \ \ \ ( n = 1,\cdots, p ).
\end{equation}
satisfies
$\ a_{i} > 0 \ \ \ (i = 1,\cdots, p)\,.$

(iv) For $\ n = p+1,\cdots, N \,,$
there are numbers $\ \alpha_{n,k} >0 \ \ \ (k = 1,\cdots, p)\ $ such that
\begin{equation*}\label{2403aaa}
   \sum_{k=1}^{p} \alpha_{n,k} \geq 1\ \ \ \text{and}\ \ \
   \bm{x}_{n} = \sum_{k=1}^{p} \alpha_{n,k} \bm{x}_{k} \,.
\end{equation*}

Furthermore, suppose a vector $\ \bm{u} = \lp u_{1},\cdots, u_{p} \rp^T\,$ satisfies
\begin{equation}\label{2403}
   \bm{x}_{n}^{T}\bm{u} \geq 1 \ \ \ ( n = 1,\cdots, N ),
\end{equation}
then
\begin{equation}\label{2404}
   u_{i} \geq a_{i} \ \ \ ( i = 1,\cdots, p ) \,.
\end{equation}

\medskip

\noindent
{\em Proof. }
From condition (iv)
it is easy to see that \eqref{2403} is equivalent to
\begin{equation*}\label{2403b}
   \bm{x}_{n}^{T}\bm{u} \geq 1 \ \ \ ( n = 1,\cdots, p ).
\end{equation*}

For $ n=1 ,$
we set
$$
    h_{1} = \frac{1}{x_{1,1}}\lp \bm{x}_{1}^{T}\bm{u} -1 \rp \geq 0
$$
and
$$
   \overline{u}_{1}= u_{1} - h_{1} \leq u_{1} \,.
$$
Denote
$\ \bm{u}_{1} = \lp \overline{u}_{1}, u_{2},\cdots, u_{p} \rp^T .$
Then
\[
    \bm{x}_{1}^{T}\bm{u}_{1}
  = \ x_{1,1}\overline{u}_{1} + x_{1,2}u_{2} + \cdots + x_{1,n}u_{n}
  = \ \bm{x}_{1}^{T}\bm{u} - \lp \bm{x}_{1}^{T}\bm{u} -1\rp = 1 .
\]
Since $ \ x_{2,1}<0 $ and $ \overline{u}_{1} \leq u_{1} $,
we have
\begin{align*}
    \bm{x}_{2}^{T}\bm{u}_{1}
     = & \ x_{2,1}\overline{u}_{1} + x_{2,2}u_{2} + \cdots + x_{2,n}u_{n} \\
  \geq & \ x_{2,1}u_{1} + x_{2,2}u_{2} + \cdots + x_{2,n}u_{n} \\
  \geq & \ \bm{x}_{2}^{T}\bm{u} \geq 1 .
\end{align*}
Now we set
$$
    h_{2} = \frac{1}{x_{2,2}}\lp \bm{x}_{2}^{T}\bm{u}_{1} -1 \rp \geq 0
$$
and
$$
    \overline{u}_{2}= u_{2} - h_{2} \leq u_{2} \,.
$$
Denote
$\ \bm{u}_{2} = \lp \overline{u}_{1}, \overline{u}_{2}, u_{3},\cdots, u_{p} \rp^T . $
Then
\begin{align*}
    \bm{x}_{2}^{T}\bm{u}_{2}
  = & \ x_{2,1}\overline{u}_{1} + x_{2,2}\overline{u}_{2} + x_{2,3}u_{3} + \cdots + x_{2,n}u_{n} \\
  = & \ \bm{x}_{2}^{T}\bm{u}_{1} - \lp \bm{x}_{2}^{T}\bm{u}_{1} - 1 \rp = 1 .
\end{align*}
Sequentially, we can define
$\
    \overline{u}_{1}, \cdots, \overline{u}_{p},
$
such that
\begin{equation}\label{2406}
    \overline{u}_{n} \leq u_{n} \ \ \ ( n = 1,\cdots, p ) .
\end{equation}
Denote
$\     \bm{u}_{p} =  \lp \overline{u}_{1}, \cdots,\overline{u}_{p} \rp^T . $
Then
$$
     \bm{x}_{n}^{T}\bm{u}_{p} = 1  \ \ \ ( n = 1,\cdots, p ).
$$
Noting that $ \bm{a} $ is unique solution of \eqref{2405aaa},
we must have $\ \bm{u}_{p} = \bm{a} ,\ $ or
$$
    \overline{u}_{n} = a_{n} \ \ \ ( n = 1,\cdots, p ) ,
$$
which combined with \eqref{2406} yields \eqref{2404} .
\eop

\bigskip

\noindent
{\em Proof of \bf Proposition 3.1. }
Suppose $ P $ is an absolutely continuous distribution over $ \real^{p\times N}$.
Let $ \mathcal{S} $
be the set of all $\ p\times N-$matrices
$ \bm{X} $
satisfying conditions (i), (ii), (iii) and (iv) in Lemma A.6.
Obviously $ \mathcal{S} $ is an open set in $ \real^{p\times N}.$
Thus $ P(\mathcal{S}) > 0$.
\eop

\bigskip

\noindent
{\em Proof of \bf Proposition 3.2. }
Denote
$$
     K = \lb \bm{a}+\bm{u}:\ \bm{u} = \lp u_{1},\cdots, u_{p} \rp^T\ \ \text{such that}\ \ a_{i}u_{i}\geq 0\ (i=1,\cdots,p) \rb\,.
$$
Without any loss of generality we may assume
$\ a_{i} > 0\ \ (i=1,\cdots, p)\,$ and then
$$
    u_{i} \geq 0\ \ (i=1,\cdots, p)\,.
$$
Clearly,
$$
    \bm{a} = \underset{\bm{w}\in K}{ \arg\min } \left\| \bm{w}\right\|^2\,.
$$
Since $\ F\subset K \,$ it is clear that $\ \bm{w} = \lp w_{1},\cdots, w_{p} \rp^T \in F\ $ implies
$$
    w_{i} \geq a_{i} > 0\ \ (i=1,\cdots, p)\,,
$$
and then
$$
  \left\| \bm{b}\odot\bm{w}\right\|^2 = b_{1}^2 w_{1}^2 + \cdots + b_{p}^2 w_{p}^2 \geq  b_{1}^2 a_{1}^2 + \cdots + b_{p}^2 a_{p}^2 \,.
$$
Thus
$$
    \bm{a} = \underset{\bm{w}^{T}\bm{x}_{n}\geq 1,\, \forall n }{ \arg\min } \left\| \bm{b}\odot\bm{w}\right\|^2 .
$$
By taking $\ \bm{b} = \bm{h}_{\infty}^{1/2} \,,$
we get $\ \widetilde{\bm{w}} =  \widehat{\bm{w}} ,$
where
$$
    \widehat{\bm{w}} = \underset{\bm{w}^{T}\bm{x}_{n}\geq 1,\, \forall n }{ \arg\min } \left\| \bm{w} \right\|^2 .
$$
Thus the asymptotic direction of GD iterates \eqref{29}, $\ \widehat{\bm{w}}\big/\|\widehat{\bm{w}}\| \,,$
is equal to $\ \widetilde{\bm{w}}\big/\|\widetilde{\bm{w}}\| ,$ which is the asymptotic direction of AdaGrad iterates \eqref{1}.
\eop

\end{document}